\documentclass[letterpaper, 10 pt, conference]{ieeeconf}  
\IEEEoverridecommandlockouts                              
\IEEEoverridecommandlockouts

\usepackage{cite}
\usepackage{amsmath,amssymb,amsfonts}
\usepackage{algorithmic}
\usepackage{graphicx}
\usepackage{textcomp}
\usepackage{xcolor}
\usepackage{balance}
\usepackage{adjustbox}
\usepackage{multirow}
\usepackage{subcaption}
 
\include{bib_ieee_abbr.def}
\include{bib_short.def}

\def\BibTeX{{\rm B\kern-.05em{\sc i\kern-.025em b}\kern-.08em
    T\kern-.1667em\lower.7ex\hbox{E}\kern-.125emX}}
\begin{document}

\title{\LARGE \bf
WorldVLM: Combining World Model Forecasting and Vision-Language Reasoning
}

\author{Stefan Englmeier\textsuperscript{*}\thanks{* These authors contributed equally to this work.}$^{1}$, Katharina Winter\textsuperscript{*}$^{1}$, Fabian B. Flohr$^{1}$
\thanks{$^{1}$Munich University of Applied Sciences, Intelligent Vehicles Lab (IVL), 80335 Munich, Germany
        {\tt\small intelligent-vehicles@hm.edu}}
}

\maketitle

\begin{abstract}
Autonomous driving systems depend on on models that can reason about high-level scene contexts and accurately predict the dynamics of their surrounding environment. Vision-Language Models (VLMs) have recently emerged as promising tools for decision-making and scene understanding, offering strong capabilities in contextual reasoning. However, their limited spatial comprehension constrains their effectiveness as end-to-end driving models. World Models (WM) internalize environmental dynamics to predict future scene evolution. Recently explored as ego-motion predictors and foundation models for autonomous driving, they represent a promising direction for addressing key challenges in the field, particularly enhancing generalization while maintaining dynamic prediction. To leverage the complementary strengths of context-based decision making and prediction, we propose WorldVLM: A hybrid architecture that unifies VLMs and WMs. In our design, the high-level VLM generates behavior commands to guide the driving WM, enabling interpretable and context-aware actions. We evaluate conditioning strategies and provide insights into the hybrid design challenges.
\end{abstract}

\section{Introduction}
Autonomous driving is a particularly challenging problem in highly dynamic environments, especially in urban areas characterized by complex traffic scenes with numerous interacting agents such as vehicles and pedestrians, as well as regions like construction zones or large intersections. Achieving safe and trustworthy driving in such scenarios requires accurate scene understanding, reliable agent forecasting, effective human interaction, and precise vehicle control.
Generative models have shown potential to enhance the generalization of autonomous driving policies to diverse long‑tail scenarios, which are often rare but safety‑critical. 

Foundational Vision‑Language Models (VLMs), trained on internet‑scale corpora beyond the driving domain, embed broad world knowledge efficiently learned from image and language space, and exhibit intrinsic reasoning abilities that support decision‑making and explainability.
While VLM-based architectures have shown remarkable results on the driving task, their pre-training on 2D image data fundamentally limits their spatial reasoning capabilities~\cite{omnidrive2024, hwang2025emma}, and phenomena such as causal confusion and covariate shift reduce their reliability for ego‑trajectory prediction~\cite{chen_end--end_2024}. 

World models (WMs) are increasingly explored as efficient simulators that generate realistic scenarios by learning complex world dynamics and physical interactions, thereby acquiring an internal model of how the environment evolves over time~\cite{ha2018world}. 
WMs aim for physical reliability by capturing scene evolution for precise frame prediction, from which, for example, high-resolution sensor data can be decoded~\cite{orbis2025}. Beyond visual forecasting, models such as LAW~\cite{li2025enhancing} serve as latent-space driving policies that predict future ego-trajectories consistent with anticipated scene dynamics. They encode the environment in a compact continuous latent space, formulated as a regression problem rather than the discrete classification paradigm of LLMs\cite{winter2025generative}, making them well suited for spatial forecasting and realistic dynamics modeling with causal reasoning. While massive general-purpose models like Nvidia Cosmos~\cite{agarwal2025cosmos} leverage vast out-of-distribution data at huge scale, specialized autonomous driving models like Orbis~\cite{orbis2025} offer greater efficiency. Yet WMs lack intrinsic reasoning and decision-making essential for safe autonomous agents.

\begin{figure}[!t]
    \centering
    \vspace{2mm}
    \includegraphics[width=0.95\columnwidth]{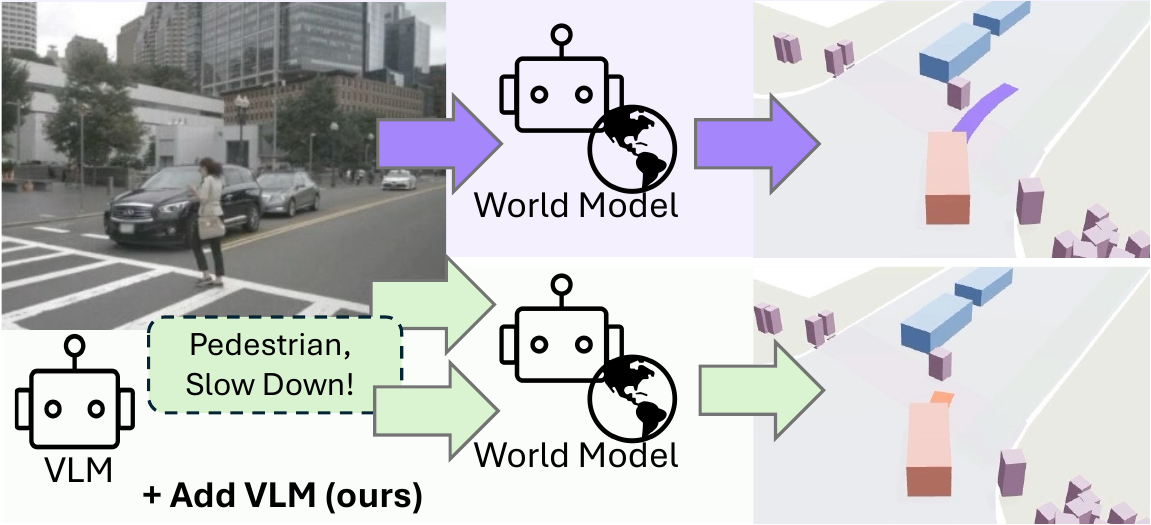} 
    \caption{We propose WorldVLM, a hybrid framework combining Vision-Language based Reasoning for high-level behavior planning and World Model forecasting for ego-trajectory prediction.}
    \label{fig:eyecatcher}
\end{figure}

\begin{figure*}[!t] 
  \centering
  \vspace{3mm}
  \includegraphics[width=0.9\textwidth]{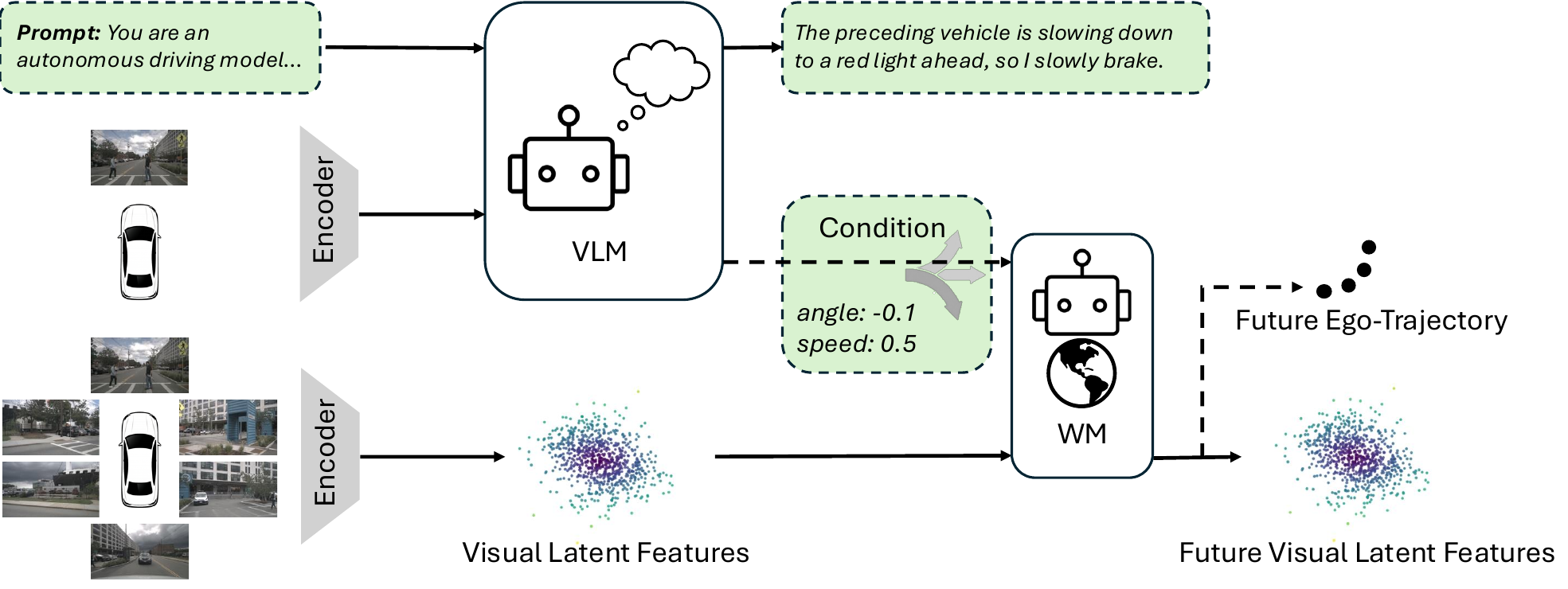} 
  \caption{WorldVLM Framework: The VLM receives front images alongside a textual prompt to generate a scene-based justification and action description and generate structured behavior commands. These commands supervise a latent driving WM~\cite{li2025enhancing} that takes visual latent features encoded from surrounding images to predict future visual latent features, extracting the ego-trajectory from the scene dynamics.}  
  \label{fig:architecture}
\end{figure*}

To address these complementary gaps, combining VLMs and WMs offers strengths for autonomous driving: reasoning and scene forecasting. Their integration for ego-trajectory planning unites high-level reasoning with realistic trajectory prediction, which can improve robustness, safety, and explainability. Similar concepts in robotics employ VLMs for task decomposition while secondary models execute behavior commands~\cite{ahn2022can, driess2023palm}. In driving, the environment consists of open world dynamics, where WMs specialize in regressing scene evolution, serving as valuable predictors executing VLM behaviors.

Our proposed model WorldVLM leverages the dynamic forecasting capabilities of WMs, conditioned by a high-level VLM that provides abstract guidance signals. The VLM generates interpretable behavior-level commands through a reasoning process for action and justification, which then condition the WM to predict the ego-trajectory, leading to safety aware trajectories (see Figure \ref{fig:eyecatcher}).

Our key contributions are twofold. 
First, we present, to the best of our knowledge, the first conceptual framework in which a VLM conditions a trajectory-predictive WM for autonomous driving through high-level behavioral commands, enabling semantically guided trajectory generation and higher interpretability of the model's intentions. We demonstrate qualitative driving scenarios and empirical experiments illustrating the feasibility and benefits of this conceptual framework in guiding behaviorally informed trajectory prediction. Second, we extend the nuScenes dataset with a justification–action annotation schema in JSON format and release all model checkpoints and source code to promote reproducibility and further research on language-guided planning.

\section{Related Work}
\subsection{Vision-Language Models in Autonomous Driving}
VLMs are increasingly adopted for trajectory planning in autonomous driving due to their strong contextual understanding, reasoning capabilities, and natural language proficiency gained from large-scale web pretraining. They can process unstructured textual prompts and produce natural-language outputs, improving explainability and facilitating human interaction. Several approaches employ Visual Question Answering (VQA)~\cite{sima2024drivelm, omnidrive2024, marcuLingoQAVisualQuestion2024} and Chain-of-Thought (CoT) reasoning~\cite{hwang2025emma, sima2024drivelm} to demonstrate high-level scene understanding. However, in end-to-end trajectory planning, VLMs exhibit challenges in spatial understanding~\cite{omnidrive2024, hwang2025emma}, particularly when precise geometry and temporally coherent motion prediction are required. To mitigate this, systems such as LMDrive~\cite{shao_lmdrive_2024} leverage multi‑modal sensor data and BEV‑decoder‑produced tokens, while BEVDriver~\cite{winter2025bevdriver} explicitly builds on a latent bird eye view (BEV) representation to strengthen spatial grounding. SimLingo~\cite{renz2025simlingo} achieves strong closed-loop performance on the CARLA Leaderboard 2.0 using a Qwen-based VLM with front-view images.
While these methods demonstrate impressive results in simulation, VLM-based systems alone do not fully exploit physically grounded inductive biases for accurate, fine-grained trajectory generation. We therefore let the VLM handle high-level behavioral reasoning and use its outputs to condition a trajectory-predictive WM that executes the underlying fine-grained path planning.

\subsection{Autonomous Driving World Models}

WMs maintain an internal representation of environmental dynamics, enabling future frame prediction~\cite{ha2018world}. In autonomous driving, they primarily serve two roles: (1) sensor generation for data augmentation, as in GAIA-2~\cite{russell2025gaia2}, Vista~\cite{gao2024vista}, Orbis~\cite{orbis2025}, and DriveDreamer4D~\cite{zhao_2025_drivedreamer4d}; and (2) latent-space forecasting, exemplified by LAW~\cite{li2025enhancing}, which encodes multi-view images into compact visual latent features or BEV maps to prioritize scene realism and spatio-temporal consistency over pixel-level sensor fidelity.
Latent-space models like LAW excel at physically coherent scene evolution and ego-trajectory prediction but lack high-level semantic reasoning for behavioral decision-making. VLM-based WMs partially address this: Hermes~\cite{zhou2025hermes} and OccLLaMA~\cite{wei2024occllama} repurpose discrete statistical token prediction for continuous frame forecasting, while ADriver-I~\cite{ADriverI2023} couples Large Language Model action selection with costly diffusion-based scene generation.
WorldVLM instead positions the VLM as a high-level behavioral planner—generating justified action commands that condition a lightweight latent WM for precise ego-trajectory regression, ensuring both semantic alignment and physical realism.

\section{Method}
Figure~\ref{fig:architecture} illustrates the architecture of WorldVLM. Given a front-view image, the VLM first generates a justification followed by an action statement, forming a structured reasoning trace that culminates in a high-level behavior command (e.g. angle and speed). Our key innovation is using this maneuver command to condition the off-the-shelf WM (LAW~\cite{li2025enhancing}). LAW takes the history of visual latent features plus the VLM's behavior token as input to predict future scene latent features. Our conditioning ensures these predictions align with the reasoned behavioral intent. The ego-trajectory is then extracted directly from these conditioned latent predictions. The VLM and conditioning mechanism are detailed below.

\subsection{Behavior-Planning VLM}
\label{subsec:behaviour}
In our proposed framework, the VLM acts as the high-level behavior planner. It receives a single front view image of the current timestep, alongside the nuScenes navigation instructions (turn left, turn right, go straight) and the current timeframe's ego-speed information.
These tokens and a numerical navigation instruction are input to the VLM following a prompt describing the task and the output schema, which contains a structured JSON with three steps: 1) justification, 2) action, 3) action token. 
The justification is a free-text reasoning describing why the selected action is safe and appropriate. The action contains a free-text natural text description of what the ego-vehicle should do next. 
The action token needs to be selected from one of the following lateral actions: \{Left Turn, Right Turn, Straight, Follow Lane, Lanechange to Left, Lanechange to Right\} paired with one of the longitudinal actions: \{Stop, Accelerate, Slow Down, Maintain\}. 

The text output is followed by a behavior command, which constitutes a structured and transparent behavior vector, offering intermediate interpretability to the overall framework. 
To generate the behavior vector, a behavior head is implemented as a three-layer multilayer perceptron with interleaved ReLU activations and dropout (dropout probability: 0.5), mapping aggregated final-layer language model hidden states to a low-dimensional control output representing a 2D steering–velocity vector. As input to this additional head, we use hidden states corresponding either to (i) a fixed subset of generated output tokens (e.g., the first 16 positions) or (ii) a set of dedicated special behavior tokens appended to the VLM output sequence.

We derive the behavior target from the ground-truth future trajectory by using the first and last points in the 3\,s horizon, $p_f=(p_{f,x},p_{f,y})$ and $p_l=(p_{l,x},p_{l,y})$. The net displacement is $\Delta p=p_l-p_f=(\Delta p_x,\Delta p_y)$ with length $d=\lVert \Delta p\rVert_2$. From this we compute a normalized velocity proxy
\[
v=\frac{d}{30},
\]
where $30$ is a normalization constant chosen so most values lie in $[0,1]$. The normalized steering target is obtained from the displacement direction
\[
\alpha=\frac{\operatorname{atan2}(\Delta p_y,\Delta p_x)+\pi}{2\pi}-0.75,
\]
where the $-0.75$ term shifts the reference direction such that the forward direction corresponds to $\alpha\approx 0$. Let the model predict $(\hat{\alpha},\hat{v})$; the behavior loss is the sum of MSE terms,
\[
\ell_{\text{behavior}}=\operatorname{MSE}(\hat{\alpha},\alpha)+\operatorname{MSE}(\hat{v},v),
\]
and the reasoning objective uses the standard token-level negative log-likelihood (cross-entropy) for next-token prediction,
\[
\ell_{\text{text}}
= -\frac{1}{T-1}\sum_{t=1}^{T-1}\log p_{\theta}\!\left(y_{t+1}\mid y_{\le t},x\right)
\]
The final objective is formulated as
\[
\ell=\ell_{\text{behavior}}+\ell_{\text{text}}.
\]

\subsection{Trajectory-Predicting World Model}

\begin{figure}[!t]  
    \vspace{2mm}
    \centering
    \includegraphics[width=0.95\columnwidth]{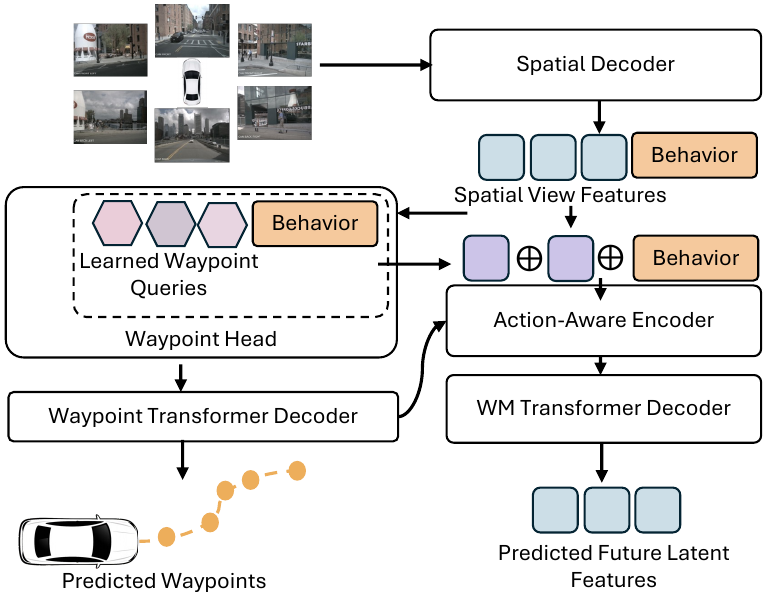}
    \caption{Behavior conditioning of the LAW~\cite{li2025enhancing} model. The behavior is concatenated with waypoint queries and spatial view features into the Waypoint Transformer Decoder and concatenated for WM prediction.}
    \label{fig:method-details}
\end{figure}

As WM, we adopt LAW~\cite{li2025enhancing}, a latent action-aware predictive model that jointly forecasts ego-centric waypoints and future visual features from multi-view image representations. LAW first extracts spatial view features via a transformer-based spatial decoder, then applies a waypoint transformer decoder to a fixed set of learned waypoint queries to obtain predicted waypoints, while a dedicated WM transformer decoder predicts the next-step latent features.

As illustrated in Figure \ref{fig:method-details}, behavior conditioning is injected at two points: the behavior vector is zero-padded to a fixed embedding size of 8 and concatenated with both the learned waypoint queries and the spatial view features before waypoint transformer decoding, and the predicted waypoints are subsequently flattened and concatenated with spatial features and behavior before being processed by an action-aware MLP and the WM transformer decoder to produce behavior-consistent future latent features. We conduct ablations in Section~\ref{subsec:ablation} comparing conditioning with and without additional behavior concatenation for scene reconstruction.

\section{Experiments}
Our modular framework allows interchangeable VLM and WM. We selected accessible, off-the-shelf models (e.g., LAW~\cite{li2025enhancing}) to demonstrate the concept's feasibility rather than pursuing state-of-the-art performance on individual modules.
Future work will explore stronger WMs (e.g., Orbis-scale models~\cite{orbis2025}) and more advanced VLMs with temporal multiview 3D grounding, alongside end-to-end training of the conditioning mechanism and closed-loop evaluation to further unlock the framework's potential.

\subsection{Setup}
For our experiments, we use LLaVA-Qwen1.5-0.5B\footnote{https://huggingface.co/IoanRazvan/LLaVA-Qwen1.5-0.5B-pretrained} and LLaVA-Qwen2-1.5B\footnote{https://huggingface.co/IoanRazvan/LLaVA-Qwen2-1.5B-pretrained} as VLM due to their compact sizes and strong reasoning capabilities and LAW~\cite{li2025enhancing} as WM.
For baseline comparison, we evaluate an unconditioned LAW variant where with "no behavior", effectively zero-padding the additional 8 behavior components..
Due to the lack of language annotations in the source VQA datasets, we create a new split from the existing train split by randomizing scenes 80\%  (22516 samples) training and 20\% validation (5615 samples). We will publish the exact split for reproducibility. We specify all dataset splits used. 
Long reasoning output cause a VLM inference time of roughly 1s, whereas LAW runs at ca. 12Hz on a NVIDIA GeForce RTX 4090 GPU.

\subsection{Dataset and Benchmark}

\begin{figure}[!t]
    \centering
    \vspace{2mm}
    \includegraphics[width=0.49\textwidth]{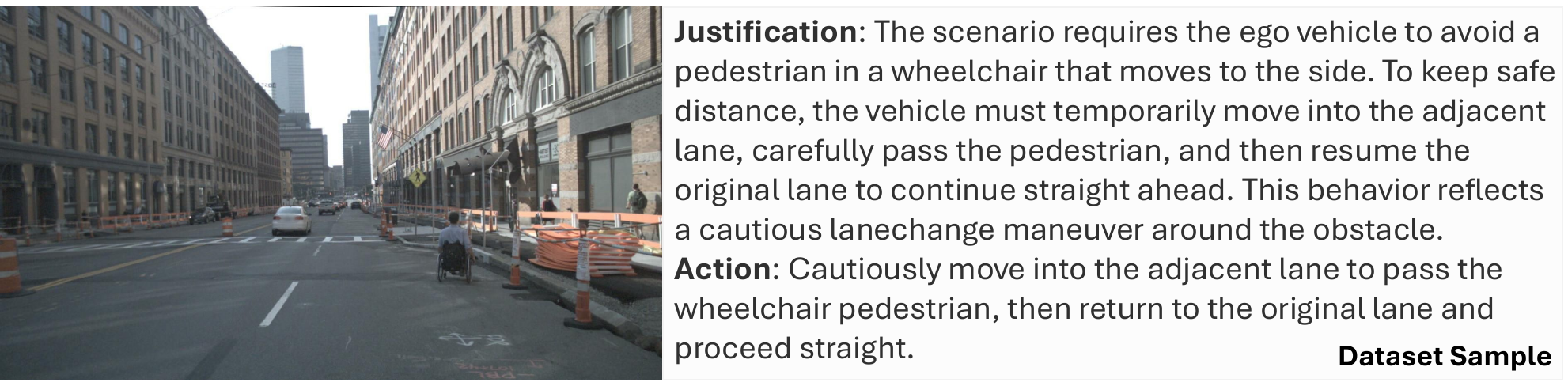} 
    \caption{Sample of our Justification and Action Description dataset.}
    \label{fig:dataset-sample}
\end{figure}

We extend nuScenes with action-justification annotations generated by combining doScenes scene-level instructions~\cite{doscenes2025itsc} and DriveLM frame-level VQA data~\cite{sima2024drivelm}. DriveLM provides QA annotations for keyframes capturing meaningful ego actions (lane changes, stops, starts) across perception, prediction, planning, and behavior categories.
These are structured into frame-level prompts fed to GPT-OSS-120B\footnote{https://ollama.com/library/gpt-oss:120b}, enforcing JSON output with justification, action, and action-enum fields. An example is presented in Figure~\ref{fig:dataset-sample}. This distills both datasets into our reasoning format of Action and Justification for explainable driving while discarding irrelevant information. We generate the ground truth annotations for the nuScenes train split, as given ground truth annotations are unavailable for other splits. 

\begin{figure}[!t]
\vspace{2mm}
    \centering
    \includegraphics[width=0.99\columnwidth]{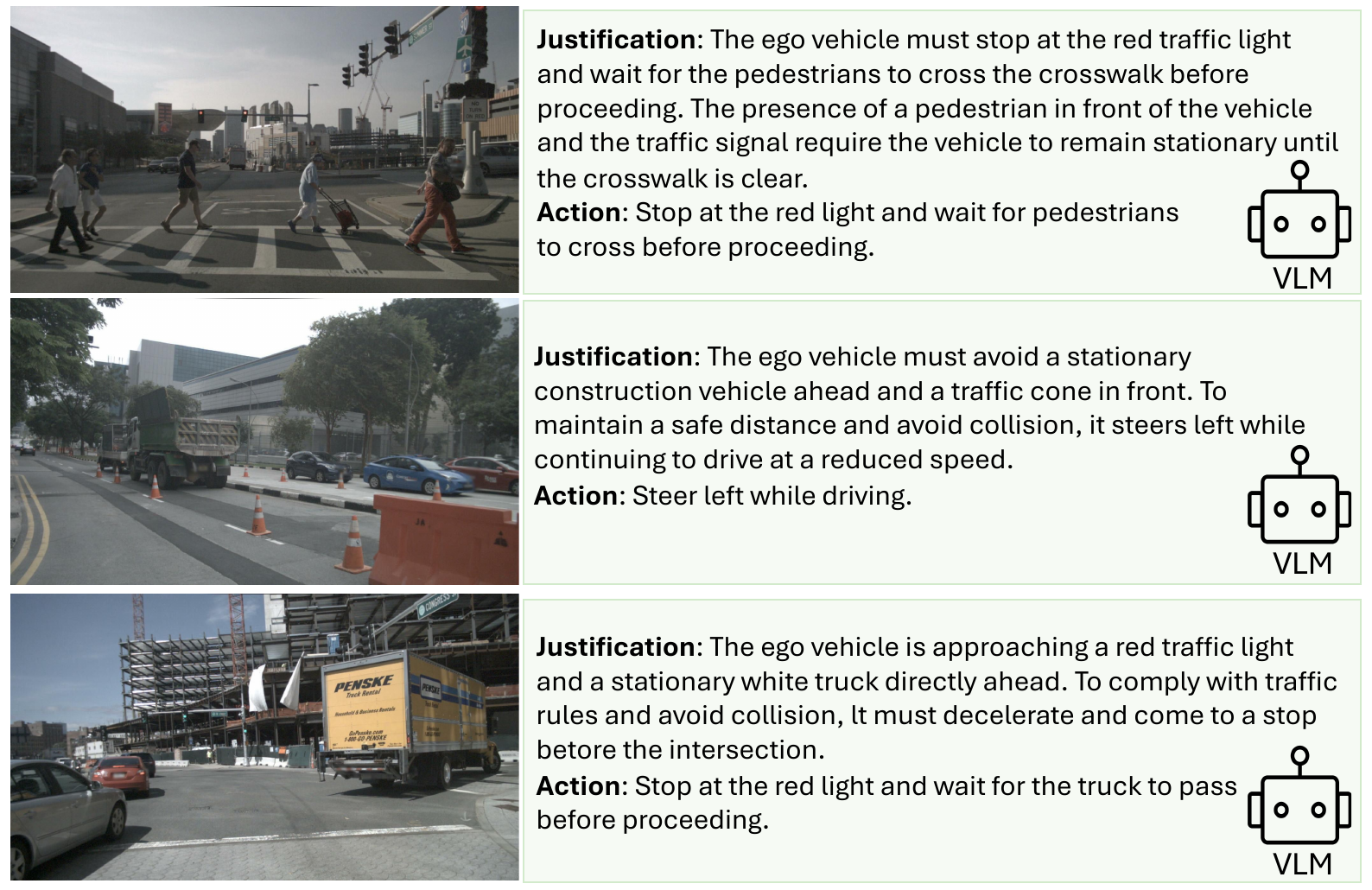} 
    \caption{Justification and Action Description generated by the behavior-planning VLM following our dataset annotation scheme.}
    \label{fig:reason-samples}
\end{figure}

\subsection{Metrics}
For open-loop evaluation, we use the nuScenes evaluation metrics L2 error in meters and collision rate in percentage.
To evaluate the quality of reasoning outputs, we use established NLP metrics BLEU~\cite{papineni2002bleu} and ROUGE~\cite{lin2004rouge}. BLEU measures n-gram precision against the ground truth from our scene description annotations, where BLEU-1 captures unigram overlap and BLEU-n (for $n>1$) combines 1- through n-gram precisions to reward both correct word choice and increasingly fluent phrase matches. ROUGE complements this with a recall; we report ROUGE-N for 1- and 2-gram overlaps and ROUGE-L for the longest common subsequence, with all scores ranging from 0 to 1. Besides these text-level metrics, BERTScore~\cite{zhang2020bert} compares token embeddings to assess semantic similarity beyond surface-form matching.

\subsection{Training}
The overall framework is trained in two stages. First, we train the VLM to generate reasoning and structured behavioral instructions that condition the WM.  Second, we train the WM on conditioning, leveraging the published vision encoder checkpoint for resource-efficient decoder training. For conditioning we use outputs of the trained VLM. For WM training, we adopt the training configurations described in LAW~\cite{li2025enhancing}. We train action and vector conditionings using four NVIDIA A40 GPUs with 46GB VRAM.
During training of the VLM, we employ the loss functions described in Subsection \ref{subsec:behaviour}. The learning rate for the VLM is initialized at $1e-6$ with $100$ warm-up steps and a weight decay of $0.06$, decaying to $1e-7$. We train the VLM for four epochs. Training is performed with a batch size of $3$ and distributed across two GPUs. For LAW training, we assume the parameters used by the authors, training on 4 GPUs.

\subsection{Qualitative Results}\label{subsec:qualitative-results}

\begin{figure*}[!htbp]
    \vspace{3mm}
    \centering
    \begin{subfigure}{0.328\textwidth}
        \centering
        \includegraphics[width=\linewidth]{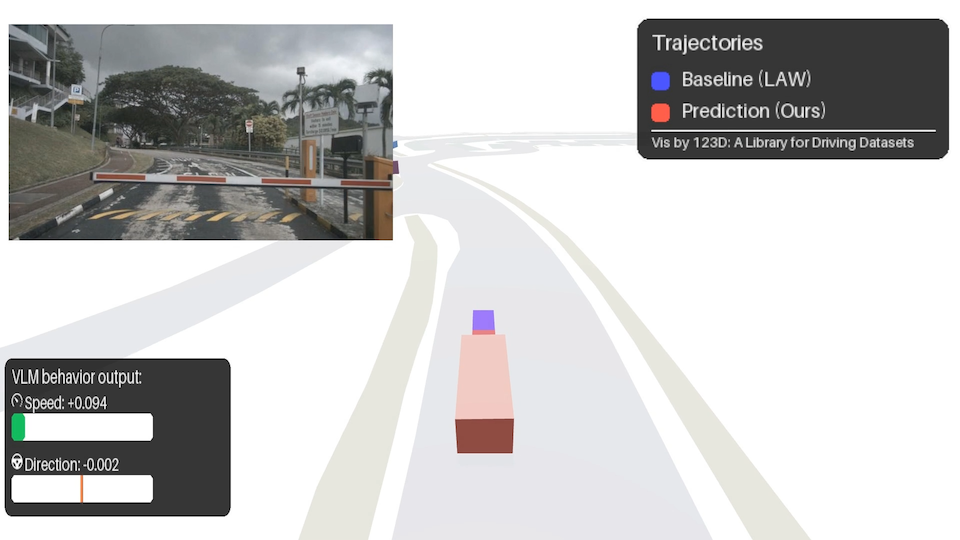}
        \caption{Gate}
        \label{fig:sub2}
    \end{subfigure}
    \begin{subfigure}{0.328\textwidth}
        \centering
        \includegraphics[width=\linewidth]{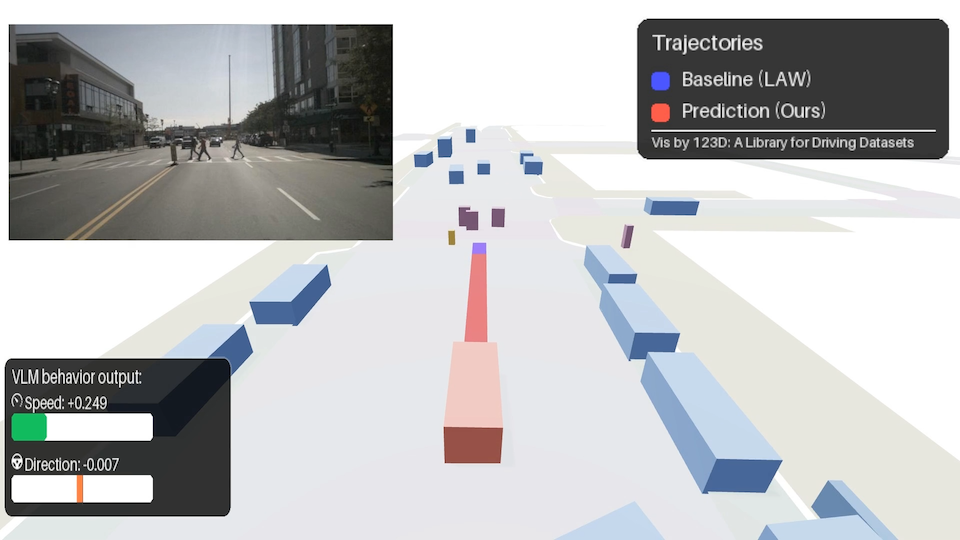}
        \caption{Pedestrians}
        \label{fig:sub1}
    \end{subfigure}
        \begin{subfigure}{0.328\textwidth}
        \centering
        \includegraphics[width=\linewidth]{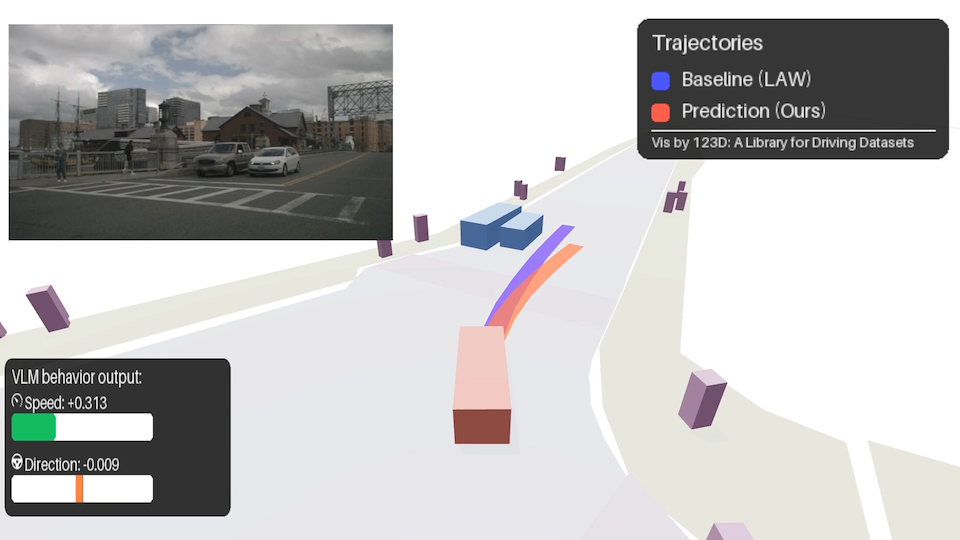}
        \caption{Right Turn}
        \label{fig:sub2}
    \end{subfigure}
    
    \vspace{0.5em}
    
    \begin{subfigure}{0.328\textwidth}
        \centering
        \includegraphics[width=\linewidth]{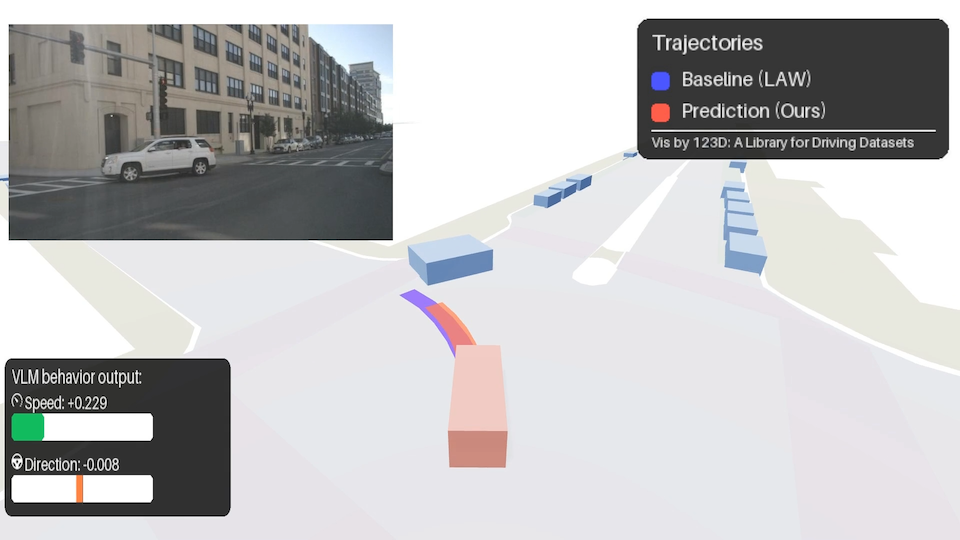}
        \caption{Left Turn Yield}
        \label{fig:sub2}
    \end{subfigure}
    \begin{subfigure}{0.328\textwidth}
        \centering
        \includegraphics[width=\linewidth]{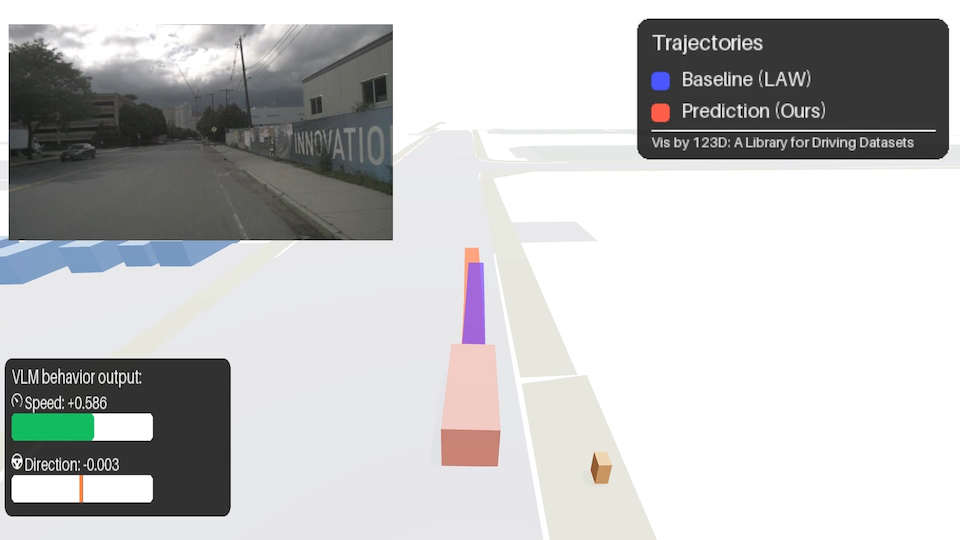}
        \caption{Empty Street}
        \label{fig:sub2}
    \end{subfigure}
    \begin{subfigure}{0.328\textwidth}
        \centering
        \includegraphics[width=\linewidth]{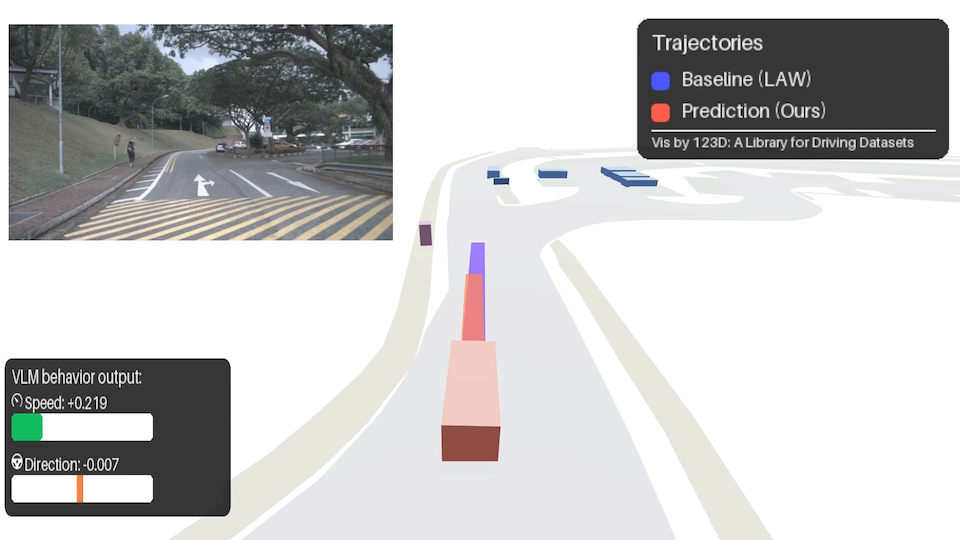}
        \caption{Speed Bump and Pedestrian}
        \label{fig:sub2}
    \end{subfigure}
        
    \caption{Qualitative comparison of safer WorldVLM (red) vs. LAW baseline (blue) ego trajectories on nuScenes validation samples. Front-view VLM input shown top left. VLM-generated direction/speed is given bottom-right. Visualized with 123D.}
    \label{fig:qualitative}

\end{figure*}

\paragraph{Reasoning}

Figure~\ref{fig:reason-samples} shows representative outputs from the fine-tuned VLM (LLaVA-Qwen1.5) trained with our justification–action dataset annotations on nuScenes.
In the first example, the model correctly detects pedestrians crossing at a busy intersection and justifies a stopping action. The second depicts a construction zone with a construction vehicle. Here, the VLM proposes a safe steering maneuver supported by accurate scene recognition.
When perception errors occur, the explicit observation–action linkage enables tracing the visual evidence underlying each decision. In the third example, the model misclassifies the traffic light as red. Although slowing down is appropriate due to a truck ahead, the justification reveals a second incorrect cause of the action.

\paragraph{Open-Loop Results}
Figure~\ref{fig:qualitative} shows WorldVLM trajectory predictions versus LAW baseline across diverse nuScenes scenarios. While aggregate metrics remain comparable, WorldVLM demonstrates more cautious behavior in interaction-heavy cases, such as maintaining greater safety margins around VRUs and other obstacles and yielding appropriately at merges and exploiting free lanes when available.

\subsection{Quantitative Results}
\paragraph{Combined Framework}
Table~\ref{tab:main-results} compares L2 error and collision rates with LAW~\cite{li2025enhancing} and our zero-padded adaptation. Our method matches baseline L2 accuracy across horizons (0.31 vs.\ 0.31, 0.62 vs.\ 0.61, 1.03 vs.\ 1.02), indicating that VLM-based conditioning preserves trajectory fidelity, but exhibits higher long-horizon collision rates. Qualitative analysis in Section~\ref{subsec:qualitative-results} shows more conservative yielding to VRUs in interactive scenarios compared to zero-padding, suggesting that VLM conditioning promotes risk-aware behavior despite aggregate metric trade-offs. While zero-padding slightly increases collision rates, this effect is mitigated at 1s and 2s by behavior conditioning.

\begin{table}[htb]
    \caption{Ground Truth runs on the nuScenes original dataset validation split. A checkmark for Nav indicates concatenation of the nuScnees navigation instruction to behavior conditioning. The angle and speed Vector is generated by the VLM.}
    \centering
    \renewcommand{\arraystretch}{1.2}
    \begin{adjustbox}{max width=\columnwidth}
    \begin{tabular}{l|c|ccc|ccc}
        \multirow{2}{*}{\textbf{Model}} & \multirow{2}{*}{\textbf{Nav}} 
        & \multicolumn{3}{c|}{\textbf{L2 (m)}} 
        & \multicolumn{3}{c}{\textbf{Collision (\%)}} \\
        &  & \textbf{1s ↓} & \textbf{2s ↓} & \textbf{3s ↓} 
           & \textbf{1s ↓} & \textbf{2s ↓} & \textbf{3s ↓} \\
        \hline
        Baseline (LAW)~\cite{li2025enhancing} & \checkmark
        & \textbf{0.31} & \textbf{0.61} & 1.02 
        & \textbf{0.10} & \textbf{0.14} & \textbf{0.44} \\
        No Behavior & \checkmark
        & \textbf{0.31} & \textbf{0.61} & \textbf{1.01} & 0.12 & 0.16 & 0.49 \\
        WorldVLM with Motion Vector & -- &
        \textbf{0.31} & 0.62 & 1.03 & \textbf{0.10} & \textbf{0.14} & 0.48 \\
    \end{tabular}
    \end{adjustbox}
    \label{tab:main-results}
\end{table}

\paragraph{Reasoning}
Table \ref{tab:nlp-metrics} reports the evaluation of the learned justification action reasoning against the dataset ground truth on our small data split. We compare BERT, ROUGE, and BLEU scores of our trained VLMs with an untrained baseline. The improvements indicate that the finetuned models learn to generate appropriate justifications and actions from the provided front-view images. All fine-tuned models achieve identical BERTScore values, yielding a 24\% improvement over zero-shot and indicating stronger semantic alignment. Fine-tuned Qwen1.5 substantially improves recall (ROUGE: 47\% vs. 9\%) and precision (BLEU: 36\% unigram and 15\% 3-gram overlap, compared to near-zero in zero-shot).
No significant differences are observed between leveraging the first 16 tokens or appending specialized behavior tokens into the behavior head for the generation of action conditions. Although the larger model underperforms the smaller variant, training dynamics suggest under-optimization rather than a fundamental limitation.

\begingroup
\setlength{\tabcolsep}{3pt} 
\begin{table}[]
\vspace{2mm}
    \caption{Evaluation of our VLM reasoning using our model with 5 specialized behavior tokens compared to the ground truth dataset on our split. *Zero-shot evaluation $^{\dagger}$ Behavior generation on the first 16 tokens.}
    \centering
    \begin{tabular}{l|ccc|ccc|ccc}

        \multirow{2}{*}{\textbf{Model}} &
        \multicolumn{3}{c|}{\textbf{ BERT (↑)}} &
        \multicolumn{3}{c|}{\textbf{ROUGE (↑)}} &
        \multicolumn{3}{c}{\textbf{BLEU (↑)}} \\
        & \textbf{F1} & \textbf{P} & \textbf{R}
        & \textbf{1} & \textbf{2} & \textbf{L}
        & \textbf{1} & \textbf{2} & \textbf{3}  \\
        \hline
        Qwen1.5-0.5B*
        & 0.54 & 0.64 & 0.52
        & 0.09 & 0.03 & 0.07
        & 0.04 & 0.02 & 0.03 \\

        Qwen1.5-0.5B
        & 0.67 & 0.68 & 0.66
        & 0.47 & 0.19 & 0.34
        & 0.36 & 0.22 & 0.15 
        \\

             Qwen1.5-0.5B$^{\dagger}$
      & 0.67 & 0.68 &  0.66
        & 0.47 & 0.19 &  0.34

        & 0.36 &  0.22 & 0.15 \\
        
        Qwen2-1.5B
        &  0.67 & 0.68 & 0.66 
        & 0.29 & 0.11 &  0.20

        & 0.17 & 0.10 & 0.06  \\

    \end{tabular}
    \label{tab:nlp-metrics}

\end{table}
\endgroup

\subsection{Ablations}\label{subsec:ablation}

\paragraph{Conditioning Architecture}

\begin{table}[!t]
    \vspace{3mm}
    \caption{Ablations with and without behavior concatenation for WM prediction on the original nuScenes datasplit. Nav is checked if navigation-instructions are additionally concatenated to behavior conditioning. Concat is checked if the behavior is additonally concatenated into the WM Transformer Decoder.}
    \centering
    \renewcommand{\arraystretch}{1.2}
    \begin{adjustbox}{max width=\columnwidth}
    \begin{tabular}{l|c|c|ccc|ccc}
        \multirow{2}{*}{\textbf{Model}} & \multirow{2}{*}{\textbf{Nav}} 
        & {\textbf{Concat}} 
        & \multicolumn{3}{c|}{\textbf{L2 (m)}} 
        & \multicolumn{3}{c}{\textbf{Collision (\%)}} \\
        & & & \textbf{1s ↓} & \textbf{2s ↓} & \textbf{3s ↓} 
           & \textbf{1s ↓} & \textbf{2s ↓} & \textbf{3s ↓} \\
        \hline
        Baseline (LAW)~\cite{li2025enhancing} & \checkmark & 
        & 0.31 & 0.61 & 1.02 
        & 0.10 & 0.14 & 0.44 \\
        \hline
        No Behavior & \checkmark & \checkmark
        & 0.31 & 0.61 & 1.01 & 0.12 & 0.16 & 0.49 \\
        Motion Vector & -- & \checkmark
        & \textbf{0.20} & \textbf{0.27 }& \textbf{0.27} & \textbf{0.07} & \textbf{0.09} & \textbf{0.10} \\
        \hline
        No Behavior & \checkmark & --
        & 0.31 & 0.61 & 1.01  
        & 0.12 & 0.16 & 0.49 \\
        Motion Vector & -- & --
        &\textbf{ 0.20 }& 0.28 & 0.28 & 0.13 & 0.11 & 0.11 \\

    \end{tabular}
    \end{adjustbox}
    \label{tab:concatenation-ablation}
\end{table}
To isolate the role of explicit intent signals, we ablate behavior concatenation in the WM head using ground truth angle and speed Motion Vectors, where future latent features are predicted solely from visuals and waypoints without additional behavior concatenation in the WM Transformer Decoder. Table~\ref{tab:concatenation-ablation} shows that L2 error remains similar in both variants improving baseline and zero-padded variant by 73\% on the 3s horizon, behavior concatenation reduces collision rate, indicating that conditioning is effective and helps disambiguate commands and avoid behavior‑agnostic futures that hurt reconstruction quality. As depicted in Figure~\ref{fig:method-details}, our main results concatenate behavior features in the WM head.

\paragraph{Conditioning Types}

\begin{table}[!htb]
    \vspace{3mm}
    \caption{Ground truth conditioning type ablations on our small datasplit. Nav is checked if navigation-instructions are additionally concatenated to behavior conditioning.}

    \centering
    \renewcommand{\arraystretch}{1.2}
    \begin{adjustbox}{max width=\columnwidth}
    \begin{tabular}{l|c|ccc|ccc}
        \multirow{2}{*}{\textbf{Model}} & \multirow{2}{*}{\textbf{Nav}} 
        & \multicolumn{3}{c|}{\textbf{L2 (m)}} 
        & \multicolumn{3}{c}{\textbf{Collision (\%)}} \\
        &  & \textbf{1s ↓} & \textbf{2s ↓} & \textbf{3s ↓} 
           & \textbf{1s ↓} & \textbf{2s ↓} & \textbf{3s ↓} \\
        \hline
       Baseline (LAW)~\cite{li2025enhancing} & \checkmark
        & 0.30 & 0.59 & 0.98 
        & 0.50 & 0.60 & 1.0 \\
       No Behavior & \checkmark
        & 0.35 & 0.67 & 1.09 & 0.10 & 0.16 & 0.51\\
         \hline
         Action (Nav x Speed) & -- &
         0.40 & 0.74 & 1.19 & 0.10 & 0.35 & 0.97 \\
         Action (Nav x Speed) & \checkmark &
         0.33 & 0.62 & 1.0 & 0.07 & 0.17 & 0.39 \\
        \hline
        Motion Vector & --
        & 0.20 & \textbf{0.27} & \textbf{0.28} & 0.06 & \textbf{0.07} & \textbf{0.10} \\
        Motion Vector & \checkmark &
        0.22 & 0.31 & 0.38 & 0.34 & 0.29 & 0.33 \\
        Motion Vector angle only & --
        & 0.37 & 0.72 & 1.18 & \textbf{0.06} & 0.19 & 0.60 \\
        Motion Vector speed only & --
        & \textbf{0.19} & 0.32 & 0.49 & 0.07 & 0.08 & 0.19 \\
        \hline
        Discrete Speed Action & --
        & 0.32 & 0.61 & 1.0 & \textbf{0.06} & 0.18 & 0.45 \\
        Discrete Speed Goal & --
        & 0.28 & 0.50 & 0.78 & 0.13 & 0.21 & 0.40 \\
    \end{tabular}
    \end{adjustbox}
    \label{tab:conditioning-ablation}
\end{table}

We study which conditioning signal best supports WM ego-trajectory prediction by comparing several behavior encodings, with results in Table~\ref{tab:conditioning-ablation}. We train and evaluate on our small reasoning datasplit, to make conditioning comparable to the annotation-derived Actions. 
We evaluate (i) Motion Vector, a continuous signal (angle, waypoint distance) with longitudinal-only and lateral-only ablations and (ii) two variants of five trajectory-derived discrete one-hot-encoded labels for acceleration instruction (Halt, Accelerate, Stop, Maintain, Decelerate) versus goal-speed bins (Halt, Very Slow, Slow, Moderate, Fast), alongside a re-trained LAW and a zero-padded “no-behavior” variant. Both conditionings are derived from the nuScenes ground truth trajectories. (iii) Action, a discrete command coming from our reasoning dataset (find details in Section \ref{subsec:behaviour}).

(i) The Motion Vector without navigation instructions achieves the best L2 (28\% on 3s) and collision rate (0.1\% on 3s), consistent with it providing the most direct, low-noise signal about the target waypoint, while framewise navigation tokens inject long-horizon noise that can mislead conditioning. Ablations show that angle-only conditioning harms performance (1.18 on 3s L2), whereas speed-only retains most of the vector’s benefits and best performance is achieved using both scalar inputs. We assume that using only angle diverts model focus.
(ii) Discrete ego-motion binning by fixed thresholds yields lower L2 error for goal-oriented than acceleration (Action) commands. Action signals reduce short-horizon collisions, while goal signals better prevent long-horizon crashes. 
Both results hold on the original nuScenes validation split.
(iii) For the reasoning data-derived Actions, navigation instructions slightly improve L2 error over the no-behavior baseline and drastically reduce collisions over Actions without navigation instructions, as coarse labels seem to benefit from added semantic context, unlike fine-grained vectors, which suffer from conflicting long-horizon bias. Despite two different sources of ground-truth, collision rates drasticlaly improve on the 3s horizon to 0.39\% compared to 1\% for baseline LAW.

Summarized, both ablations indicate that goal-oriented signals excel in L2 error and long-horizon collision avoidance, while action commands better reduce short-horizon crashes. Signals from diverse data sources improve both metrics despite slight L2 divergence. Fine-grained commands boost open-loop performance with accurate inputs.

\paragraph{Action Token Representation}
Table \ref{tab:token_rep} compares different token representations of our VLM used as input to the behavior head predicting speed and steering angle conditionings. We report the mean absolute error on the original data split.

We evaluate three categories: (1) using $n$ dedicated behavior tokens appended to the reasoning output, (2) using the first $n$ tokens, and (3) using the last $n$ tokens as input to the head. The number of tokens is not critical when using behavior tokens; however, relying on the last-layer tokens degrades performance. Using the first $n$ tokens yields the best results, although restricting this to only the first 8 tokens is insufficient. Furthermore, there is almost no performance difference between the 0.5B and 1.5B models.
\begin{table}[!t]
\vspace{3mm}
\caption{Comparison of VLM token representations (behavior tokens, first $n$ tokens, last $n$ tokens) as input to the behavior head predicting speed and steering conditionings, evaluated using mean absolute error on the original data split.}
\vspace{5mm}
    \centering
    \renewcommand{\arraystretch}{1.2}
    \begin{adjustbox}{max width=0.6\textwidth}
    \begin{tabular}{l|c|c}
        \textbf{Method}

        & \textbf{MAE angle ↓} 
        & \textbf{MAE speed ↓} \\
 
        \hline

        1 BEH Token

          & 0.0463 & 0.1920 \\

        5 BEH Token
         & 0.0408 & 0.1992 \\

        10 BEH Token
  &  0.0416 & 0.1977 \\

                   Last 5
   & 0.0508 & 0.5768 \\
   
        Last 16
   & 0.1257 & 0.2663 \\

        First 8
   & 0.0152 & 0.2566 \\
        
        First 16
  & 0.0135 & \textbf{0.1788} \\

        First 16 1.5B
  & \textbf{0.0126} & 0.1810 \\

        First 32
   & 0.0165 & 0.2030 \\

        \end{tabular}
    \end{adjustbox}
    \label{tab:token_rep}
\end{table}

\section{Discussion}
In this work, we introduced and evaluated a new framework that combines VLMs and WMs yielding valuable insights while highlighting clear directions for improvement. 

We integrate VLM-derived behavior tokens into the WM, achieving stable L2 parity with the original baseline. While revealing long-horizon challenges, our ablations and qualitative analysis provide design guidelines for multimodal trajectory prediction. Future work targets refined VLM supervision incorporating our insights for interaction-heavy scenarios.

One limiting factor is the current single frame processing of front‑view images, which are cropped by LLaVA’s vision encoder, leading to a narrow field of view and limited spatial and temporal awareness. Future work will connect the VLM directly to the WM encoder’s latent representations, unifying perception and decision‑making in a shared latent space.

Although our experimental results show that WorldVLM produces plausible, interpretable scene‑level reasoning and qualitatively safe trajectories in complex driving scenarios, we acknowledge that the VLM‑generated reasoning dataset lacks reliable human‑annotated reasoning traces. 

In future work, we plan to close these gaps by training and evaluating the system in closed‑loop simulations with violation-sensitive safety metrics, refining conditioning with human‑annotated traces, and strengthening the bidirectional interaction between VLM and WM. For real-time feasibility, we plan to decouple inference rates by running the VLM at a lower frequency for high-level scene reasoning, while maintaining high-frequency world model updates for control and trajectory prediction. These steps are expected to improve both interpretability and performance, advancing toward robust autonomous driving grounded in high‑level scene understanding. 

\section{Conclusion}
In this work, we proposed WorldVLM, a hybrid framework that integrates a vision‑language model (VLM) generating scene reasoning into high‑level behavior commands with a low‑level world model (WM) that produces concrete motion predictions. We explored effective practices for connecting VLM outputs to latent world‑model representations and highlighted key challenges in such hybrid architectures.

WorldVLM demonstrates that grounding low‑level planning in explicit, language‑informed reasoning is feasible and promising. We believe that closing the gap between qualitative promise and quantitative performance, through improved conditioning, human‑annotated reasoning traces, and closed‑loop evaluation, will make vision‑language world models a cornerstone of interpretable and robust autonomous driving systems.

\section{Acknowledgement}
The research leading to these results is funded by the
German Federal Ministry for Economic Affairs and Energy within the project “NXT GEN AI METHODS – Generative Methoden für Perzeption, Prädiktion und Planung”
(grant no. 19A23914M) and the Federal Ministry of Research, Technology and Space (BMFTR) within
the project ADRIVE-GPT (grant no. 13FH544KA2). The authors are solely
responsible for the content of this publication.

\balance
\bibliographystyle{ieeetr}
\bibliography{references.bib}

\end{document}